\documentclass[sigconf]{acmart}
\AtBeginDocument{%
  \providecommand\BibTeX{{%
    \normalfont B\kern-0.5em{\scshape i\kern-0.25em b}\kern-0.8em\TeX}}}

\copyrightyear{2023}
\acmYear{2023}
\setcopyright{acmlicensed}\acmConference[MAD '23]{2nd ACM International Workshop on Multimedia AI against Disinformation }{June 12, 2023}{Thessaloniki, Greece}
\acmBooktitle{2nd ACM International Workshop on Multimedia AI against Disinformation (MAD '23), June 12, 2023, Thessaloniki, Greece}
\acmPrice{15.00}
\acmDOI{10.1145/3592572.3592845}
\acmISBN{979-8-4007-0187-0/23/06}

\begin{document}

\title{Examining European Press Coverage \\ of the Covid-19 No-Vax Movement: An NLP Framework}

\author{David Alonso del Barrio}
\email{ddbarrio@idiap.ch}
\affiliation{%
  \institution{Idiap Research Institute}
  \country{Switzerland}
}

\author{Daniel Gatica-Perez}
\email{gatica@idiap.ch}
\affiliation{%
  \institution{Idiap Research Institute and EPFL}
  \country{Switzerland}
}

\begin{abstract}
\footnote{\textbf{David Alonso del Barrio, Daniel Gatica-Perez| ACM 2023. This is the author's version of the work. It is posted here for your personal use. Not for redistribution. The definitive Version of Record will be published in https://doi.org/10.1145/3592572.3592845.}}
This paper examines how the European press dealt with the no-vax reactions against the Covid-19 vaccine and the dis- and misinformation associated with this movement. Using a curated dataset of 1786 articles from 19 European newspapers on the anti-vaccine movement over a period of 22 months in 2020-2021, we used Natural Language Processing techniques including topic modeling, sentiment analysis, semantic relationship with word embeddings, political analysis, named entity recognition, and semantic networks, to understand the specific role of the European traditional press in the disinformation ecosystem. The results of this multi-angle analysis demonstrate that the European well-established press actively opposed a variety of hoaxes mainly spread on social media, and was critical of the anti-vax trend, regardless of the political orientation of the newspaper. This confirms the relevance of studying the role of high-quality press in the disinformation ecosystem.

\end{abstract}




\maketitle

\section{Introduction}
The no-vax Covid-19 movement
has been driven by the ease of news dissemination, mainly through social media \cite{burki2020online}, where rumors and conspiracy theories can contribute to vaccine hesitancy \cite{islam2021covid} \cite{freeman2022covid}. 
 Both traditional and social media information sources have played a fundamental role in informing the general public about vaccination, and such information has an influence on public opinion, either by increasing doubts about the vaccine or by promoting vaccination. \cite{chen2022descriptive}.
The sources of information used by the general public in different countries on the subject of Covid-19 vaccination have been studied through surveys \cite{fletcher2018measuring} \cite{dupuis2021misinformation} \cite{piltch2021examining}. In all of them, a similar conclusion is reached, namely that people who follow more traditional media such as newspapers or TV news, have a greater predisposition to the vaccine, while people who use social media as first source of information are more reluctant to vaccination \cite{chen2022descriptive}.

Much of the existing research on the subject has focused on the detection or analysis of false content, whether text, images, or videos in social media \cite{patwa2021fighting} \cite{pennycook2020fighting}. In comparison, there has been less analysis of how high-quality press outlets dealt with the no-vax and dis- and misinformation issues \cite{semeraro2022writing}, even though they play a very important role in society.  
In terms of trust, many readers tend to rely much more on media with a proven and long journalistic tradition \cite{murphy2021psychological}.

Our work addresses this research gap by posing two research questions:

{\bf RQ1}: How did the major European newspapers present the news about the movement against the Covid-19 vaccine?

{\bf RQ2}: How did these major European newspapers deal with the dis/misinformation about the Covid-19 vaccine?

By addressing the above research questions, our work makes the following contributions:

{\bf Contribution 1}: Using a dataset of European articles on the anti-vax movement, we first carried out a series of analyses such as subtopic modeling, semantic relationships with word-embeddings, classification of articles by country, and sentiment analysis based on political orientation.

{\bf Contribution 2}: We deepened the first analysis by studying and analyzing sentences with the term disinformation (or related) in these articles, combined with other analysis techniques (named entity recognition, keyword association, sentiment analysis, and relationship between entities), which together help reflect how the European press dealt with this issue, actively playing a fundamental role against no-vax related disinformation.

The paper is organized as follows. In Section \ref{section:related_work}, we discuss related work. In Section \ref{section:data}, we present the dataset we used. In Section \ref{section:methodology}, we describe the methodology used to analyze the no-vax news data. We present and discuss results in Section \ref{section:Results_and_discussion}. Finally, we provide  conclusions in Section \ref{section:conclusions}.

\section{Related work} \label{section:related_work}


From the NLP point of view, different types of analysis have been done on social media to characterize dis/misinformation on Twitter, Facebook, Youtube, etc. Similar analyses can be applied to more traditional media, such as newspapers.

More specifically, Gao et al.\cite{gao2021changes} presented a semantic network and sentiment analysis on Weibo texts about the vaccination against Covid-19. 
Stella et al.\cite{stella2022cognitive} did similar work with tweets about Covid-19 vaccines in Italian and English.
Chen et al.\cite{luo2021exploring} did semantic network analysis on US tweets and Weibo texts in China; the results demonstrate that the two countries’ social media users overlapped in themes concerning domestic vaccination policies, priority groups, challenges from Covid-19 variants, and the global pandemic situation. However, Twitter users were prone to disclose individual vaccination experiences and to express anti-vaccine attitudes. In comparison, Weibo users manifested evident deference to authorities and exhibited more positive feelings toward the Covid-19 vaccine.
Ma et al.\cite{ma2021use} tested two methods to detect subtopics on tweets about vaccine hesitancy: LDA and Top2Vec, identifying topics such as alternative Covid-19 treatments or side effects of vaccines. Similar work was done by Jiang et al.\cite{jiang2021characterization} using LDA on Covid-19 Vaccine US Tweets, and detected seven major themes: "News Related to Coronavirus and Vaccine Development", "General Discussion and Seeking of Information on Coronavirus", "Financial Concerns", "Venting Negative Emotions", "Prayers and Calls for Positivity", "Efficacy of Vaccine and Treatment", and "Conspiracies about Coronavirus and Its Vaccines."

Examining recent work analyzing the press role, more specifically newspapers discussing the Covid-19 vaccination topic, Semeraro et al.\cite{semeraro2022writing} collected 5745 news about Covid-19 vaccines in Italy, covering 17 outlets over 8 months. The work used cognitive network science and natural language processing to reconstruct time-evolving semantic and emotional frames in Italian press.

Looking more specifically at the topic of dis/misinformation about Covid-19 in newspapers,
Rovetta et al.\cite{rovetta2021influence} made an analysis of Covid-19 headlines in Italy and the role of the mainstream press in the infodemic. Similarly, Catalan et al.\cite{catalan2020vaccine} made an analysis of the Spanish press. Both of these works follow a journalistic approach.
Our work is related to this type of work, but from a different perspective, focused on NLP. The computational approach could validate, reaffirm,  or contrast
conclusions reached from the journalistic discipline, as well as providing the possibility of comparing results across multiple European countries, thus going beyond the single-nation information ecosystem.

\section{Data} \label{section:data}
Our starting point has been a previous work \cite{alonso2022did}, where we created a dataset of more than 50,000 articles on Covid-19 vaccination with articles from Italy (2 newspapers), France (6 newspapers), Spain (6 newspapers), Switzerland (3 newspapers) and the United Kingdom (2 newspapers), with all the content translated to English.

In this previous work, sentiment analysis was performed at headline level, full article level, and sentence level, as well as subtopic detection, in order to detect sub-issues within the topic of covid vaccination. The no-vax movement was among the discovered sub-themes using BERTopic \cite{grootendorst2022bertopic} and this subtopic was defined by the following words: "anti-vaxxers", "anti-vaccine","anti-vaxx", "anti-corona", "no-vax", "no vax" and "anti-vaccin" . Our work focuses on the study and analysis of this sample of articles on the no-vax movement, which consists of 1786 articles (3.48\% of the full corpus.) 

\begin{figure}[h]
  \centering
  \includegraphics[width=\linewidth]{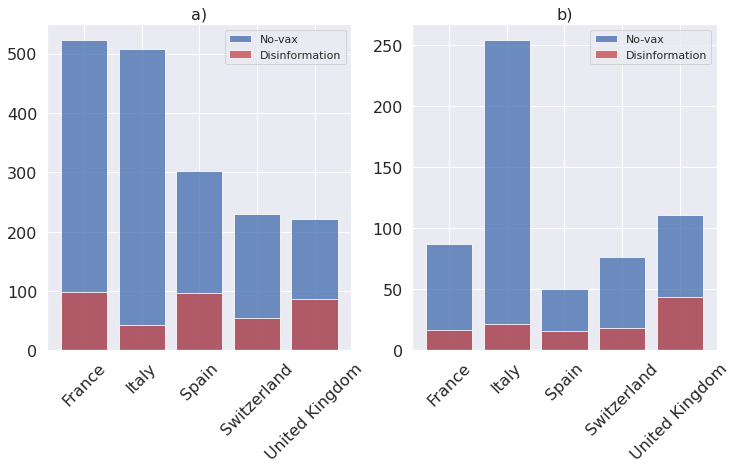}
  \caption{a)Per-country number of articles about disinformation in the dataset of articles about no-Vax. b)Per-country number of articles about disinformation in the dataset of articles about no-Vax normalized by the number of newspapers.}
  \label{fig:dis_nov}
\end{figure}

Within the no-vax theme, which is related to disinformation, as much of the cause of vaccine refusal is generated by fake news, rumors and conspiracy theories, we first set it up to obtain basic statistics to better define the dataset used. As part of the the articles on no-vax, we extracted the number of articles containing any of the following words: disinformation, misinformation, hoaxes, rumours, and theories; this filtering resulted in 380 articles (21.2\% of the 1786 articles). The results per country are shown in Figure \ref{fig:dis_nov} in the plot on the left. We see that France and Italy are the countries with the most content on no-vax. However, if we focus on the highest number of articles containing words related to disinformation, it is the United Kingdom with 39\% and Spain with 31\% of articles on disinformation within anti-vax-related news. In the plot on the right we wanted to normalize the results of the number of articles per country, normalizing by the number of newspapers per country, trying to solve the question if in Italy and France there is more No-vax content because there are more newspapers. Looking at the results we can see that Italy and UK, where there are only two newspapers, is where in relation to the number of newspapers there are more articles about No-Vax and disinformation.

\section{Methodology} 
\label{section:methodology}
In this section, we describe the different kinds of analysis we have done with the corpus.

\subsection{Subtopic modeling}
Within the issue of no-vax, we wanted to identify what other sub-themes were being addressed, to understand more clearly how the European press dealt with this issue. We used BERTopic for this purpose \cite{grootendorst2020bertopic}. BERTopic is a popular technique that uses transformers and c-TF-IDF to create clusters in which each topic is represented by a list of words, so that the interpretation of topics in most cases is  intuitive.

\subsection{Semantic similarity with word embeddings}
The idea behind the use of word embedding is to try to identify semantic relationships between the words in the articles. In this way, we intend to capture associations between different aspects of the non-vax movement. We used Word2Vec \cite{mikolov2013efficient}, a standard word embedding technique that allows to map words into a high-dimensional vector space, where semantically similar words are located close to each other.

We used the Gensim library as it has the word2vec module \cite{rehurek2011gensim}. We used very generic parameters; from the results, we consider that they were adequate. 
\begin{itemize}
\item Vector size: This parameter controls the dimensionality of the word embeddings. In this case, a vector size of 100 was chosen, which means that each word in the vocabulary will be represented by a 100-dimension vector. This is a common choice for word embedding models, as it is often sufficient to capture the semantic and syntactic properties of words.
\item Window size: This parameter determines the context window size for each word in the text. A window size of 5 was chosen, which means that the model will consider the 5 words before and after each target word in the text. This is a reasonable choice for capturing local context and is often used as a default parameter value.
\item Minimum word count: This parameter specifies the minimum frequency of a word in the corpus required to be included in the vocabulary. A value of 1 was chosen, which means that all words in the corpus will be included in the vocabulary. This is appropriate if the dataset is small or if there are many rare words in the text. We have a total of 2237630 words and 124072 different words. 
\item Number of workers: This parameter specifies the number of worker threads to use when training the model. A value of 4 was chosen, which means that 4 threads will be used to parallelize the training process. This is a reasonable choice for most machines. 
\end{itemize}
To visualize the results, we used 
t-SNE (t-Distributed Stochastic Neighbor Embedding) \cite{JMLR:v9:vandermaaten08a}, to reduce the dimensionality of the word embeddings from 100 dimensions to 2. Each word is defined as a point, and t-SNE works by first computing pairwise distances between the high-dimensional points, and then mapping these points to a lower-dimensional space while preserving the distances as much as possible.

\subsection{Classification by country}
Continuing with the aim of understanding how similar the articles on the same subject are in different European countries, we proposed the following analysis. 

We have used a dataset consisting of three columns, x, x' and y, where x is the main text of the articles, x' which is the text of the article without Named Entities and y which is the country of origin of the article. Following a classification approach, we wanted to know how easy it is to classify articles by country, and how similar are the articles between countries. The assumption is that if we remove the Named Entities (named entities are extracted using the module described in Section 4.5.3), for example, the name of cities or politicians, the texts between countries will be similar and difficult to classify. To contrast this assumption we have used Logistic Regression as a classification method, using the cross validation technique, where we have divided the dataset xy and x'y into 5 parts, where in each iteration one of the parts was used for testing, and the other 4 parts for training. In order to enter the text as input to the Logistic Regression classifier, we have removed stopwords and converted the text into vectors using TF-IDF.
Logistic regression allows us to determine the importance of features by analyzing the coefficients of the model. These coefficients' size and sign indicate the strength and direction of the relationship between each feature and the target variable. When the coefficient is positive, it means that an increase in the feature leads to an increase in the target variable's probability of being positive, while a negative coefficient implies the opposite. 
When dealing with multiclass classification, logistic regression generates a binary classifier for each class, with unique coefficients assigned to each classifier. This way, we can utilize the feature coefficients for each binary classifier to assess the significance of features for each corresponding class. So we can extract which feautures (words) are the ones that help to classify the country of the article.
\subsection{Political orientation}
Another relevant analysis is to understand possible differences in the treatment of information on this issue, depending on the political orientation of the newspapers. This is especially important as in certain countries the anti-vaccine movement has been associated with  right-wing ideologies \cite{cadeddu2020beliefs}, and thus it is interesting to test this, i.e., how newspapers of different ideologies have treated this.

For this purpose, we classified each of the 19 newspapers in the dataset, into three possible subgroups: left, right, and other. The classification has been made on the basis of the description of these newspapers in their corresponding Wikipedia page.  Those newspapers classified in Wikipedia as center-left and left were labeled in our work as 'left', those newspapers classified as center-right and right were labeled as 'right', and the rest of newspapers, defined as center, socioliberal, and liberal in Wikipedia, were labeled as 'others'. This resulted in 7 newspapers labeled as right, 4 newspapers labeled as left, and 8 newspapers labeled as others.

Using the results of the sentiment analysis carried out at article level, sentence level, and headline level in our previous work \cite{alonso2022did}, we were able to compute the majority sentiment for each political orientation. So that each headline has a sentiment. At the article level, for each article the sentiment of every sentence has been extracted and the percentage of positive, negative and neutral sentences contained in that article has been calculated. Finally, at the sentence level, what is done is that, for each sentence within an article, which contains one of the words that define the No-vax topic, the sentiment is analyzed and the percentage of positive, negative and neutral sentences in each article is calculated. We then grouped these percentages by political orientation and calculated the average percentages for each political group.

\subsection{Analysis of sentences related to disinformation}
We aimed to study in more detail those sentences that mention the word disinformation, misinformation, hoaxes, rumours, or theories, to better understand how the European press has dealt with these issues. In total, we found 805 sentences, which is enough material to make a descriptive analysis. Based on this, we conducted a series of experiments to characterize the type of dis/misinformation sentences in the European press.

\subsubsection{Sentiment analysis}
Again, we used the results of sentiment analysis at sentence level, to understand the predominant tone in the sentences about disinformation, and to contrast whether the traditional press has used a more critical tone against disinformation than social media.
\subsubsection{Keyword Association}
This experiment aimed to reveal what other words tend to appear near the 'disinformation' keyword.  This is useful for discovering patterns and trends in the data, and could be used to extract insights and generate hypotheses about the topics and themes present in the text. In the context of disinformation, keyword association can help identify words and phrases that are commonly associated with this topic, and understand the patterns of language used in the press. 
For this purpose, we used the word co-occurrence matrix, which creates a matrix that tracks the number of times each word co-occurs with a target keywords. For each sentence, it counts the number of times each word appears within the window size of the target keyword. In our case we used a window size of two, so we considered the two words before and two words after the keyword in the dataset. Then, we visualized the co-occurrence matrix with a word cloud. Word clouds are a way of visualizing the relative frequency of words in a text corpus, with the most frequent words appearing in larger sizes.

\subsubsection{Named Entity Recognition (NER)}
We used NER to identify and extract named entities (people, organizations, locations, etc.) mentioned in the articles. This helped identify key actors and groups involved in the no-vax movement, and examine their role in the dissemination of disinformation.  For this task, we  used the Spacy library, \cite{spacy2}  which is a free, open-source library for advanced Natural Language Processing (NLP) in Python.

\subsubsection{Semantic Network}
We created a semantic network of named entities, in order to understand the relationship between entities. For this we have used the Python Networkx library \cite{hagberg2008exploring}.

\section{Results and discussion} 
\label{section:Results_and_discussion}

We now present the results for each of the elements of the methodology discussed in the previous section.

\subsection{Subtopic modeling}

Table \ref{tab:subtopics} shows the sub-themes identified using BERTopic. Some of the most recurrent topics are the {\bf measures adopted by different governments}, taking into account the number of cases and the {\bf clinical situation} of the country, which makes the anti-vaccine movement react and motivates {\bf demonstrations}, and on the other hand, the impact of measures such as mandatory vaccination for different {\bf jobs}. Moreover, news about the role of {\bf social media} in spreading false news, and the measures taken by these tech companies against disinformation, as well as the position of the {\bf Catholic Church} regarding vaccination, or {\bf Brazil} and the figure of Jair Bolsonaro.
Finally, another of the sub-themes identified has been {\bf sports}, where very relevant sports figures have opposed vaccination.
\begin{table}[!htp]\centering
\caption{Identified subtopics}\label{tab:subtopics}
\scriptsize \resizebox{8.5cm}{!}{
\begin{tabular}{|l|p{5cm}|}\toprule
SUBTOPIC &KEY WORDS \\\midrule
government measures  &president, government, campaign, minister, state, authority, country, risk \\ \hline
clinical effect of vaccine &disease, virus, risk, effect, dose, hospital, tiem, study, adtum, patient \\ \hline
Demostration & demostration, demostrator, saturday, pass, police, freedom, city, place, rally, procession \\ \hline
Social Media & Facebook, network, content, platform, Twitter, misinformation, video, Youtube, information, pandemic \\ \hline
Work &vax, law, worker, court, suspension, order, decree, measure, doctor, nurse \\ \hline
Vatican & bishop, church, Pope, Francis, cardinal, Vatican, priest, faith, catholics, God \\  \hline
Brazil & Paulo, coronavac, Brazil, butantan, hair, Bolsonaro, sinovac, brazilians, doria, death \\  \hline
Sport &NBA, player, league, game, Kyrie, team, Irving, season, basketball, athlete \\
\bottomrule
\end{tabular}
}
\end{table}

\subsection{Semantic similarity with word embeddings}

\begin{figure}[b]
  \centering
  \includegraphics[width=\linewidth]{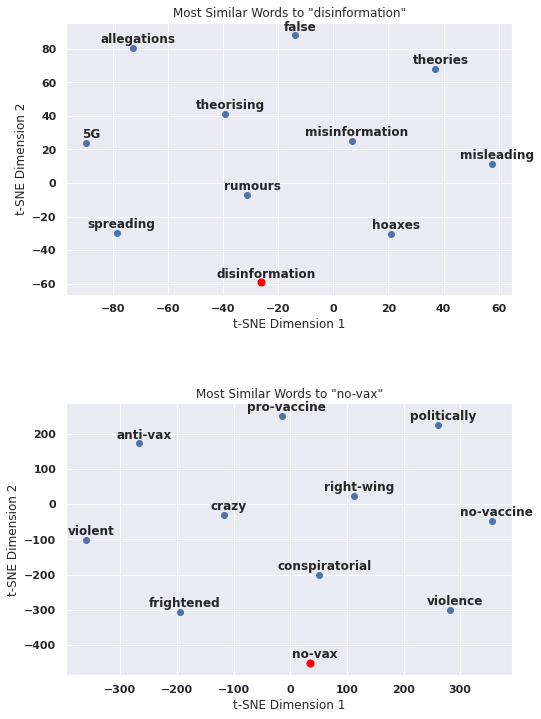}
  \caption{Word embeddings: no-vax and disinformation.}
  \label{fig:word_embeddig}
\end{figure}

Figure \ref{fig:word_embeddig} shows the semantic similarity of words that appear in the articles with the concepts of disinformation and no-vax. In the case of disinformation, we see terms such as misinformation, hoaxes, theories, false, misleading, but also 5G, since there were countless rumors about the relationship between 5G and the Covid-19 vaccine.
On the other hand, in the case of no-vax, the terms with semantic similarity are anti-vax, no-vaccine, conspiratorial, violence, but also terms such as crazy and right-wing stand out. Regarding the term crazy, it appears in sentences where doctors or professionals opine about the rumors against the vaccine, or the denial of Covid-19. With respect to the term right-wing, much of the disinformation is associated with groups like QAnon 
or figures like Bolsonaro in Brazil, who were referents of Covid-19 denial and are associated with the political right. 
\subsection{Classification by country}
As mentioned in the methodology, with this experiment we aimed to understand how different articles are across countries from the perspective of classification, using logistic regression and 5-fold cross-validation. This approach was followed for both the text with named entities and the text without named entities. The average classification accuracy in both cases was very similar, with 82\% accuracy with named entities versus 79\% accuracy in the case of articles without named entities.

\begin{figure}
  \centering
  \begin{tabular}{@{}c@{}}
    \includegraphics[width=\linewidth]{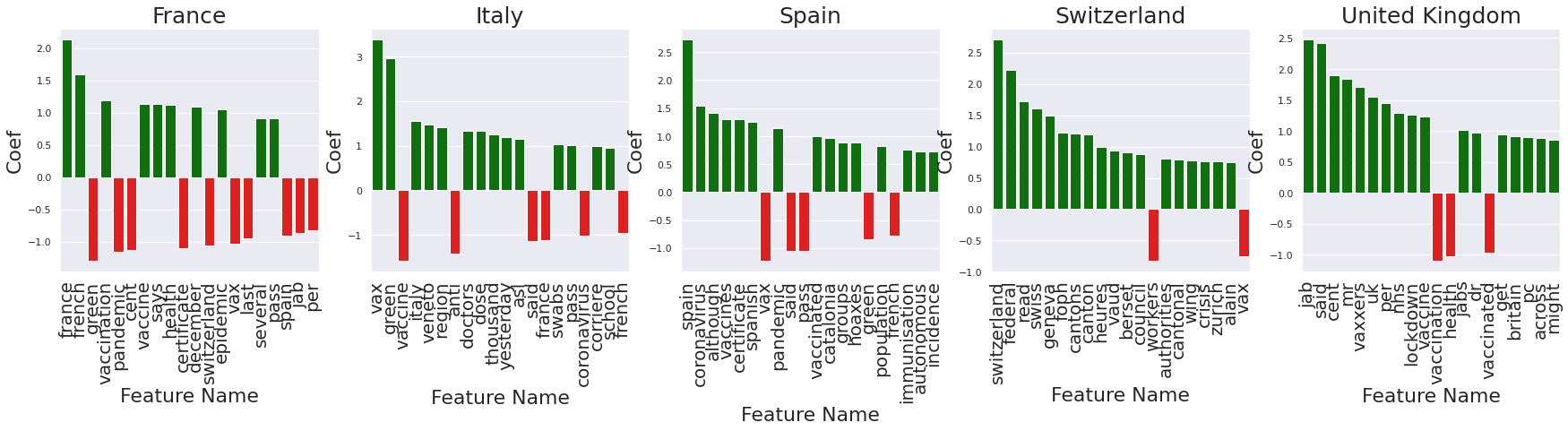} \\[\abovecaptionskip]
    \small (a) Results per country, articles With Named Entities
  \end{tabular}

  \vspace{\floatsep}

  \begin{tabular}{@{}c@{}}
    \includegraphics[width=\linewidth,scale=2]{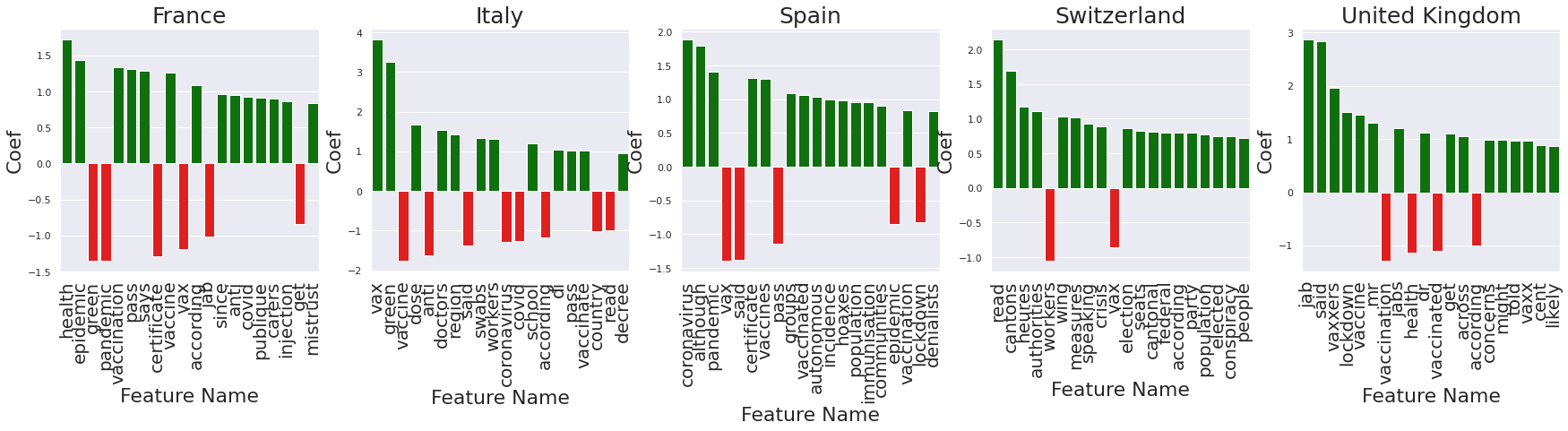} \\[\abovecaptionskip]
    \small (b) Results per country, articles Without Named Entities
  \end{tabular}

\caption{Features of Logistic regression}
  \label{fig:lr}
\end{figure}

In Figure \ref{fig:lr}, we show in green the features (words) that help classify the articles on the no-vax by country, and in red those features that do not help classify each country.  In Fig. \ref{fig:lr}(a) we take into account named entities, and in Fig. \ref{fig:lr}(b) we remove all named entities. With named entities, we see that among the main features that help classify by country is the name of the country itself, and the gentilice of the country. However, after removing all named entities, logistic regression still works quite well. This is due to specific characteristics of each country. For example, Spain has autonomous communities, while  Switzerland has cantons. As a second example, the vaccination certificate in Italy was called "green pass", while in Spain it was "Covid-19 certificate". These particularities in language make that even after removing named entities, the country that originates the news can be distinguished.  
Based on these results, we conclude that despite dealing with the same topic in all countries, and even removing named entities such as the mention of the country itself, cities, authorities, etc., the origin of the article is still distinguishable, and that reflects the locality in the language and the use of its own terms and expressions, despite having all the content translated into English.

\subsection{Political orientation}
Table \ref{tab:politics} shows the average percentage of negative sentiment as well as the negative/positive sentiment ratio at the article level, at the headline level, and at the sentence level, according to newspaper political orientation (left, right, and other). Ratios where the negative content is 5 times higher than the positive content have been marked in red.
At article and headline levels, we see that the left has a much higher ratio than the right, while at sentence level, it is the right that has a much higher ratio.

\begin{table}[!htp]\centering
\caption{Sentiment analysis by political orientation}\label{tab:politics}
\scriptsize
\resizebox{8.5cm}{!}{
\begin{tabular}{|c|c|c|c|c|c|}\toprule
\textbf{NO VAX} &\textbf{LEFT} &\textbf{ANOTHER} &\textbf{RIGHT} \\\midrule
\textbf{FREQUENCY BY IDEOLOGY; (\%)} &270 (15.1) &909 (50.9) &607 (34.0) \\
\textbf{NEG(\%);NEG/POS ARTICLE RATIO} &\textbf{29.3;\textcolor{red}{8.6}} &26.5;\textcolor{red}{6.1} &27.3;\textcolor{red}{5.5} \\
\textbf{NEG(\%);NEG/POS HEADLINE RATIO} &\textbf{37;\textcolor{red}{92.5}} &33.4;\textcolor{red}{16.7} &34.9; \textcolor{red}{23.2} \\
\textbf{NEG(\%);NEG/POS SENTENCE RATIO} & 51.4;\textcolor{red}{64.2} &47.9;\textcolor{red}{53.2} &\textbf{52.6;\textcolor{red}{105.2}}\\ 
\bottomrule
\end{tabular}
}
\end{table}

These results, where the left seems more negative than the right, fit with what was discussed in previous literature where the no-vax movement has been associated more with the right \cite{winter2022conspiracy}. It should also be noted that all the newspapers have a negative and critical tone towards this movement, which also coincides with previous research that found that   people who read the traditional press were more open to receiving the vaccine \cite{chen2022descriptive}.

\subsection{Analysis of sentences of disinformation}
\subsubsection{Sentiment analysis}
\begin{figure}[h]
  \centering
  \includegraphics[width=\linewidth]{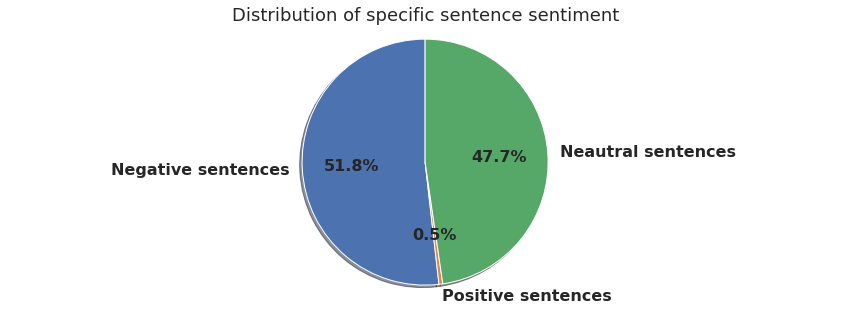}
  \caption{Sentiment analysis of disinformation sentences.}
  \label{fig:sentiment_dis}
\end{figure}

In Figure \ref{fig:sentiment_dis}, we see how more than half of the sentences containing terms related to disinformation have been labeled as negative, while only 0.5\% have been labeled positive; the rest of the cases have been labeled as neutral. Table \ref{tab:sentient_sample} shows some examples of sentences labeled as positive and negative.

\begin{table}[!htp]
    \centering
    \caption{Sample sentences about disinformation labeled by sentiment}\label{tab:sentient_sample}
    \scriptsize
    \resizebox{8.5cm}{!}{
        \begin{tabular}{|p{6cm}|c|c|}\toprule
            SENTENCES & LABELED SENTIMENT \\ \midrule
            "I spend a lot of time trying to combat misinformation and I decided to post both because I'm so proud of Phoebe and because its important to show science will win.” & POSITIVE \\ \hline
            "For the time being, the WHO is committed to combating misinformation with scientifically proven data." & POSITIVE \\ \hline
            "Clearly, our country has not yet found a way to effectively combat this disturbing misinformation." & NEGATIVE \\ \hline
            "Moreover, by fuelling conspiracy theories, it weakens our democracies." & NEGATIVE \\ \bottomrule
        \end{tabular}
    }
\end{table}

This predominantly negative tone is reflected as a criticism of the dissemination of disinformation (mainly in social media), and is in line with what has been argued in previous literature, which highlights the work of newspapers in reporting on the vaccine, rather than spreading rumors as may be happening in social media \cite{piltch2021examining}.

\subsubsection{Keyword Association}
\begin{figure}[h]
  \centering
  \includegraphics[width=\linewidth]{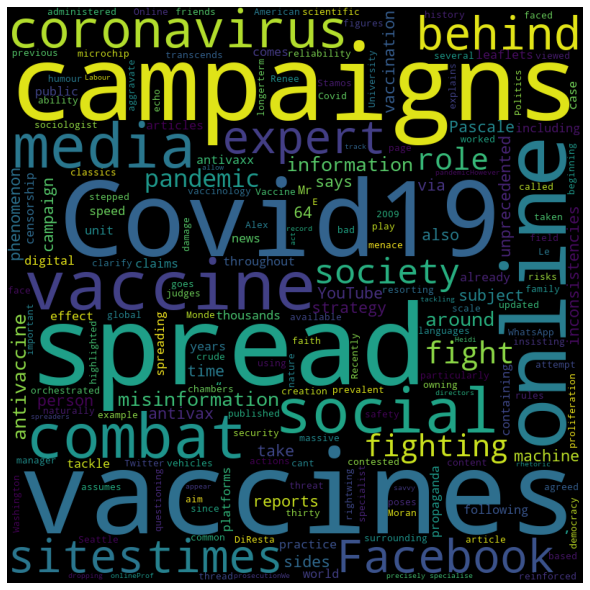}
  \caption{Wordcloud of words next to the disinformation keyword .}
  \label{fig:wordcloud_dis}
\end{figure}

Figure \ref{fig:wordcloud_dis} shows a wordcloud of the words that appear next to the word disinformation. We can highlight coronavirus, campaigns, social media, online, and spread. 
Named social networks like Facebook are also present, given their role in the spread of disinformation, rumors, and false theories.

\subsubsection{Named Entity Recognition}

\begin{figure}[h]
  \centering
  \includegraphics[width=\linewidth]{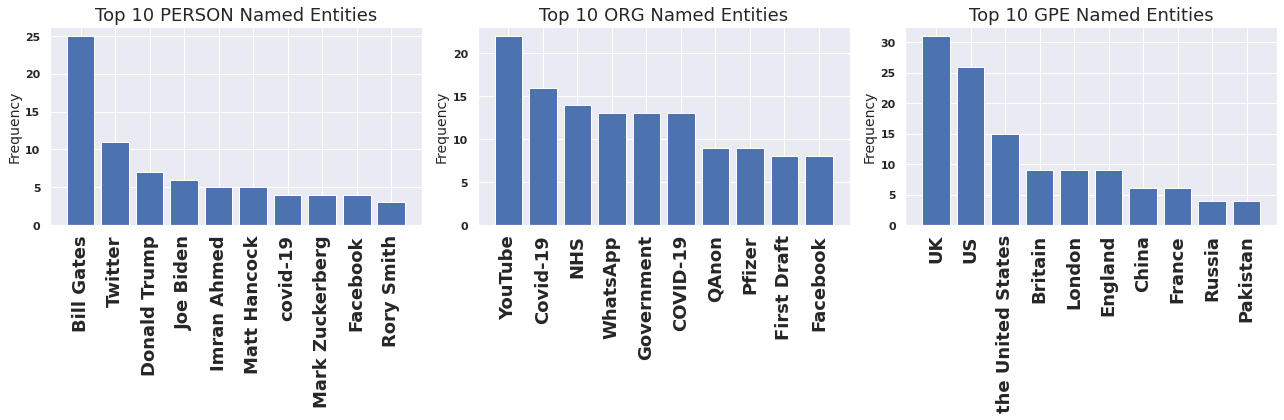}
  \caption{Most frequent named entities in disinformation sentences.}
  \label{fig:ner_dis}
\end{figure}

Figure \ref{fig:ner_dis} shows the top 10 people, organizations, and geopolitical entities (GPE) most frequently mentioned in the sentences about disinformation.

\begin{itemize}
    \item Person: One name stands out above the rest, Bill Gates, who has been the protagonist of many conspiracy theories. He is followed by the last two U.S. presidents, Donald Trump and Joe Biden: the former, strongly associated with the QAnon movement, known for its conspiracy theories, and the latter, named for his criticism of social networks for spreading disinformation. 
     \item Organizations: Social media such as Facebook, Youtube, Instagram, WhatsApp and QAnon have been heavily criticized for spreading disinformation.
     \item GPE: United Kingdom and the United States are the most mentioned countries
\end{itemize}

\subsubsection{Semantic Network}
\begin{figure}[h]
  \centering
  \includegraphics[width=\linewidth]{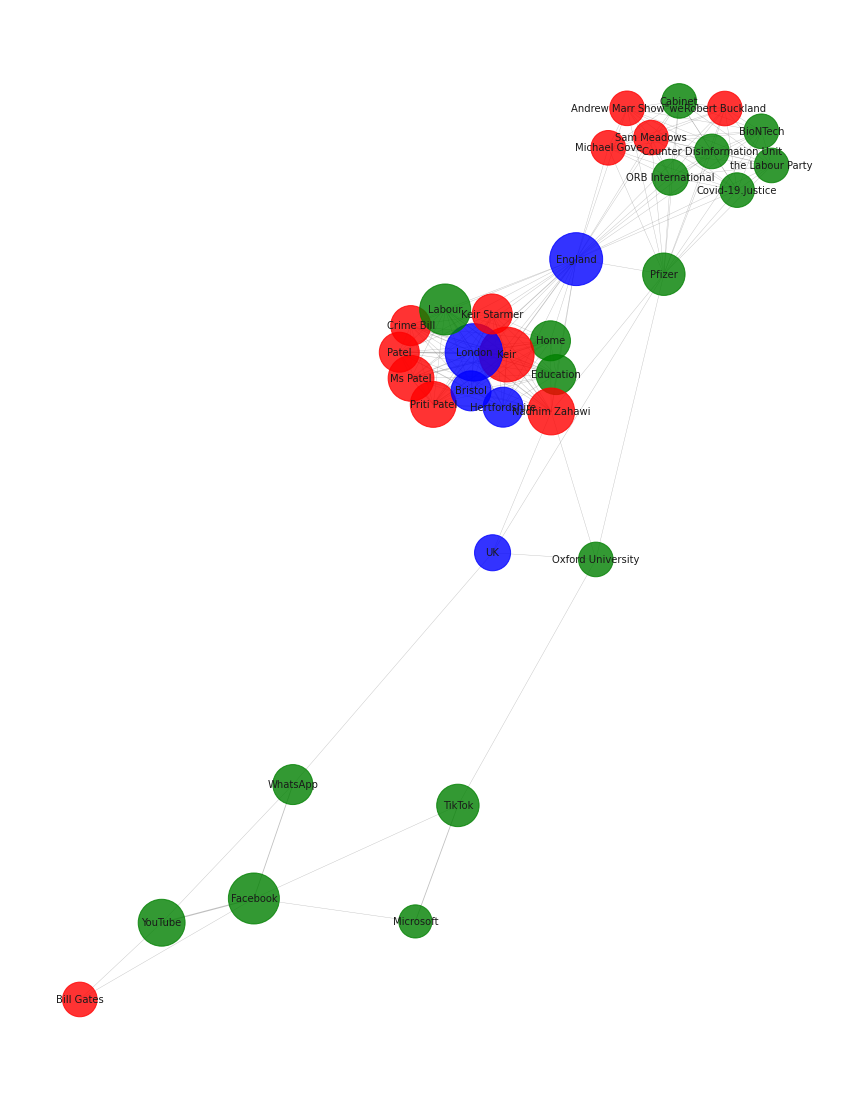}
  \caption{Semantic Network of Named Entities.}
  \label{fig:network_dis}
\end{figure}

Based on the previous identification of named entities, Figure \ref{fig:network_dis} shows the relationship among them through a semantic network. Each node is a named entity, and its size depends on the number of times it is named (the more it is named the bigger the node is.) We have also assigned colors depending on the type of entity. People are represented in red, organisations in green, and countries/cities in blue. We  weighted edges based on the frequency of co-occurrence of the entities in the sentences so that the edges are slightly thicker depending on the frequency, and we  used a spring layout instead of a circular layout for the nodes, which help to separate the nodes and edges more clearly. We set a minimum of 10 appearances, because otherwise there were too many nodes. Even with these adjustments, there were too many nodes, and the associations between named entities could not be seen correctly, so we decided to filter the input dataset. We chose only sentences containing the word disinformation (105 items) instead of taking those sentences plus sentences containing misinformation, hoaxes, rumours, and theories. Therefore, the result of the figure is based on 105 sentences containing the word "disinformation".
In this Figure, we see three big clusters. Starting from the bottom, we see Bill Gates related to social networks: Youtube, Facebook, etc. These social networks are related to the United Kingdom (second block) where we see political figures such as Patel, Starmer and Zahawi who have been very critical of social networks and the dissemination of disinformation. Above is another group made up mainly of organisations, such as Pfizer, BioNTech, the Labour Party, ORB International, and Counter Disinformation Unit, who appear in sentences about the race for vaccine development and the effect it has on the spread of disinformation.

\subsection{Comparison with related work in social media}

It is important to contrast our results with those of related work on the analysis of social media and disinformation, to better understand the entire ecosystem where dis/misinformation is present \cite{jallen}.

With regard to the detected sub-themes, we can identify some relationships between themes. For example, in the work of Nuzhath et al. \cite{nuzhath2020covid} the topic of religious beliefs is associated with  our topic of Vatican, where in "some churches and Christian ministries with large online followings — as well as Christian influencers on Facebook, Instagram, TikTok, Twitter and YouTube — are making false claims that vaccines contain fetal tissue or microchips, or are construing associations between vaccine ingredients and the devil. Others talk about how coronavirus vaccines and masks contain or herald the “mark of the beast,” a reference to an apocalyptic passage from the Book of Revelation that suggests that the Antichrist will test Christians by asking them to put a mark on their bodies" \cite{dwoskin_2021}.
Other similarities between themes include "vaccine   safety and effectiveness" with our topic "clinical effect of vaccine", and  "freedom of choice" related to our "demonstration" and "government measures" topics, as people took to the streets to express their opposition to the compulsory measures and to claim their right to choose. 
If we now take a look at works that have analyzed political orientation in social media and the press about Covid-19 vaccination, Sharma et al.\cite{sharma2022covid} present some conclusions that we have paraphrased: 
\begin{itemize}
    \item The anti-vaccine and far-right communities have a greater influence on right-leaning communities, as seen by the difference in vaccination rates between left and right states in the US.
    \item Although they produce fewer tweets overall, anti-vaccine misinformation communities are the most active in discussions about vaccines.
    \item Far-right conspiracy groups are more vocal in non-vaccine discussions and are more connected with accounts that have more followers and followings. In vaccine discussions, mainstream news and left-leaning communities are more active than far-right and right-leaning ones, but less active than the anti-vaccine community.
\end{itemize}
The last reported finding by Sharma et al.\cite{sharma2022covid} about left-wing news being more vocal, perhaps could be associated it in our case with the greater amount of negative tone we found in headlines generated by left-oriented newspapers in Europe.
 

Finally, if we look in more detail at the misinformation and conspiracies about Covid-19 both in the news and on social media, Shahsavari et al. \cite{shahsavari2020conspiracy} presents arguments that are largely in line with our findings. They state that
"Among the stories currently circulating in US-focused social media forums are ones suggesting that the 5G telecommunication network activates the virus, that the pandemic is a hoax perpetrated by a global cabal, that the virus is a bio-weapon released deliberately by the Chinese, or that Bill Gates is using it as cover to launch a broad vaccination program to facilitate a global surveillance regime" \cite{shahsavari2020conspiracy}. All this is related with our findings where Bill Gates is one of the most frequent named entities, or the presence of 5G as a term semantically closed to the word disinformation. 

In short, we observe that the European press played an active role in disproving conspiracies and false information, which had their breeding ground mainly in social media \cite{allington2021health}.

\section{Conclusions} \label{section:conclusions}

In this paper, based on news data from 19 European newspapers, 5 countries, and 22 months, we first presented an NLP-based, multi-faceted analysis of news articles about the no-vax movement in Europe, and then we examined the issue of disinformation with a fine-grain analysis at sentence level. Overall, our work contributes to the understanding of how mis- and disinformation is talked about in the larger information ecosystem, including high-quality, long-standing media outlets.

We conclude by answering our two research questions:

RQ1: How did the major European newspapers present the news about the movement against the Covid-19 vaccine?
After analyzing the corpus on the no-vax movement, we can say that the press has played the role of reporting on events such as demonstrations and government action. We can also say that there is a general negative tone towards the anti-vaccine movement, regardless of the political orientation of the newspaper, as well as a great locality in the vocabulary used by the newspapers in different countries, which makes it relatively easy to identify the country of origin of the information.

RQ2: How did these major European newspapers deal with the dis/misinformation about the Covid-19 vaccine?
After analyzing the sentences in the corpus related to  disinformation in the context of the Covid-19 vaccine, we  highlight the negative tone of the press on this issue, playing a fundamental role in disproving disinformation often published on social media, as well as unfounded theories such as the relation of the 5G or Bill Gates with the vaccine.

Overall, the availability of a longitudinal, multi-country, multi-outlet news corpus enabled the study presented here. Future work could extend the analysis by systematically and quantitatively comparing the tone, used language, and argumentation used by high-quality newspapers and social media outlets on the same specific disinformation-related events or campaigns, to examine how strongly different the positions can be in practice.







\begin{acks}
This work was supported by the AI4Media project, funded by the European
Commission (Grant 951911) under the H2020 Programme ICT-48-2020.
We also thank the newspapers that agreed to share their online articles. 
\end{acks}

\bibliographystyle{ACM-Reference-Format}
\bibliography{sample-base}


\end{document}